\documentclass{article}

\usepackage{PRIMEarxiv}

\usepackage[utf8]{inputenc} % allow utf-8 input
\usepackage[T1]{fontenc}    % use 8-bit T1 fonts
\usepackage[hidelinks]{hyperref}
\usepackage{url}            % simple URL typesetting
\usepackage{booktabs}       % professional-quality tables
\usepackage{amsfonts}       % blackboard math symbols
\usepackage{nicefrac}       % compact symbols for 1/2, etc.
\usepackage{microtype}      % microtypography
\usepackage{lipsum}
\usepackage{fancyhdr}       % header
\usepackage{graphicx}       % graphics
\graphicspath{{media/}}     % organize your images and other figures under media/ folder
\usepackage{amsmath}
\usepackage[usestackEOL]{stackengine}
\usepackage{multirow}

\title{VNIbCReg: VICReg with Neighboring-Invariance and better-Covariance Evaluated on Non-stationary Seismic Signal Time Series
\thanks{This paper is a part of the following paper: \cite{lee2022ensemble}. The proposed self-supervised learning part is taken from \cite{lee2022ensemble} so that this paper would be more accessible and readable for audience in machine learning.}
}

\author{
  Daesoo Lee \\
  Norwegian University of Science and Technology \\
  \texttt{daesoo.lee@ntnu.no} \\
   \And
  Erlend Aune \\
  Norwegian University of Science and Technology, \\ BI Norwegian Business School \\
  \And
  Nad\`ege Langet \\
  Norwegian Seismic Array \\
  \And
  Jo Eidsvik \\
  Norwegian University of Science and Technology \\
}

\begin{document}
\maketitle

\begin{abstract}
One of the latest self-supervised learning (SSL) methods, VICReg, showed a great performance both in the linear evaluation and the fine-tuning evaluation. However, VICReg is proposed in computer vision and it learns by pulling representations of random crops of an image while maintaining the representation space by the variance and covariance loss. However, VICReg would be ineffective on non-stationary time series where different parts/crops of input should be differently encoded to consider the non-stationarity. Another recent SSL proposal, Temporal Neighborhood Coding (TNC) is effective for encoding non-stationary time series. This study shows that a combination of a VICReg-style method and TNC is very effective for SSL on non-stationary time series, where a non-stationary seismic signal time series is used as an evaluation dataset. 
\end{abstract}

% keywords can be removed

\section{Introduction}

The recent mainstream SSL frameworks can be divided into two main categories: 1) contrastive learning, 2) non-contrastive learning. Some well-known contrastive learning methods are MoCo \cite{he2020momentum} and SimCLR \cite{chen2020simple}. In those methods, there are a reference sample, a positive sample, and a negative sample. The reference and positive samples form a positive pair, and the reference and negative samples form a negative pair. Then, those contrastive methods learn representations by pulling the representations of the positive pairs together and pushing those of the negative pairs apart. However, these methods require a large number of negative pairs per positive pair to learn representations effectively. To eliminate the need for negative pairs, non-contrastive learning methods such as BYOL \cite{grill2020bootstrap}, SimSiam \cite{chen2021exploring}, Barlow Twins \cite{zbontar2021barlow}, VICReg \cite{bardes2021vicreg}, and VIbCReg \cite{lee2021vibcreg} have been proposed. The non-contrastive learning methods use positive pairs only, and training of the networks could be simplified. The non-contrastive learning methods were able to outperform the existing contrastive learning methods in terms of quality of learned representations. In particular, VICReg and VIbCReg outperform other competing SSL methods by having effective feature decorrelation. 

Temporal Neighborhood Coding (TNC) \cite{tonekaboni2021unsupervised} is another recent SSL method to learn representations for non-stationary time series. It learns time series representations by ensuring that a distribution of signals from the same neighborhood is distinguishable from a distribution of non-neighboring signals. It was developed to address time series in the medical field, where modeling the dynamic nature of time series data is important. 

\begin{figure}[htb]
    \centering
    \includegraphics[width=1.0\textwidth]{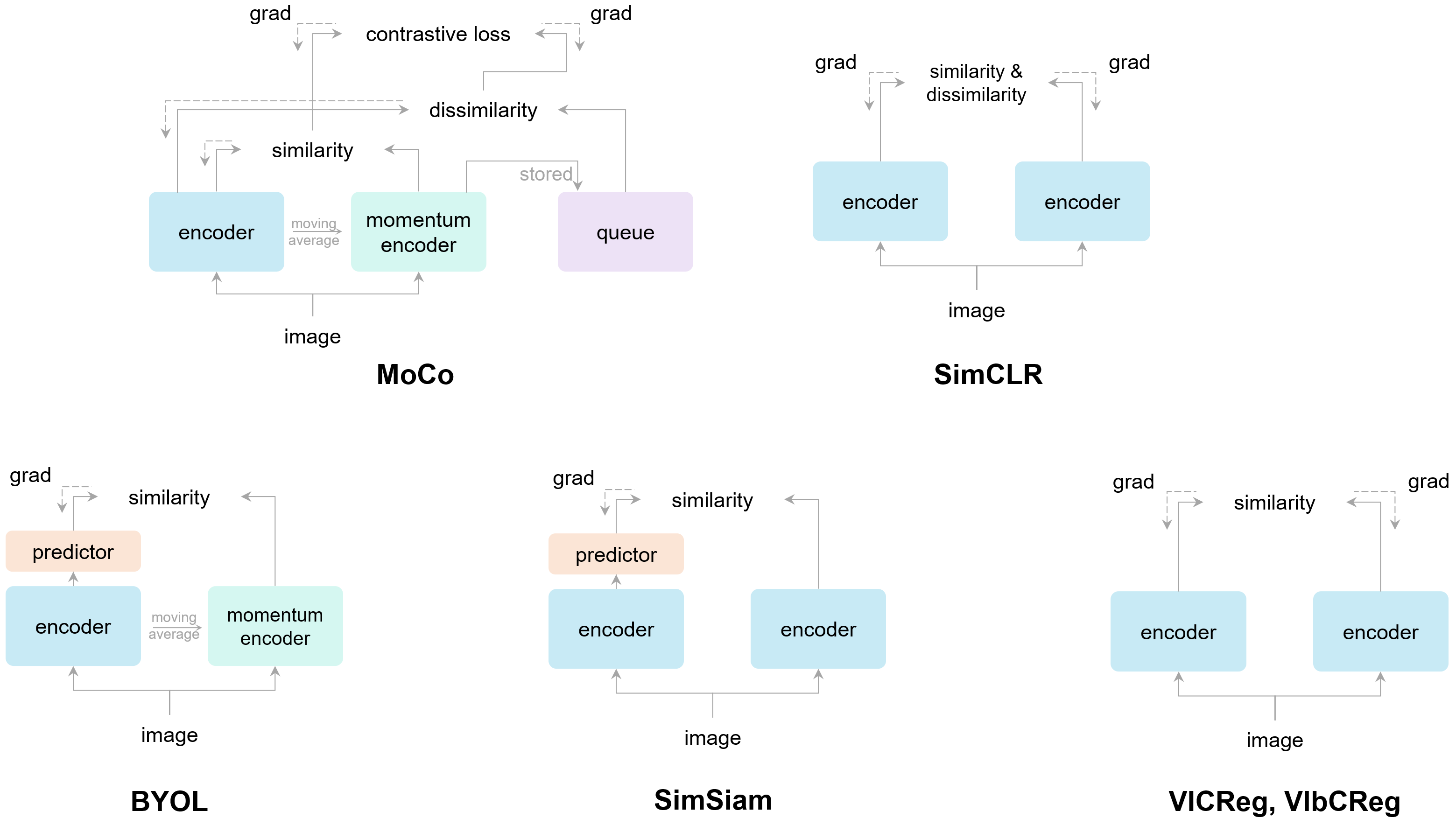}
    \caption{Simplified illustrations of SSL methods. grad denotes gradient and similarity denotes similarity between two output vectors from two encoders that share their weights and the similarity is optimized to be reduced between the two augmented images. The dash lines indicate the gradient propagation flow. Therefore, the lack of a dash line denotes stop-gradient.}
    \label{fig:ssl_methods}       % Give a unique label
\end{figure}

\begin{figure}[htb]
    \centering
    \includegraphics[width=1.0\textwidth]{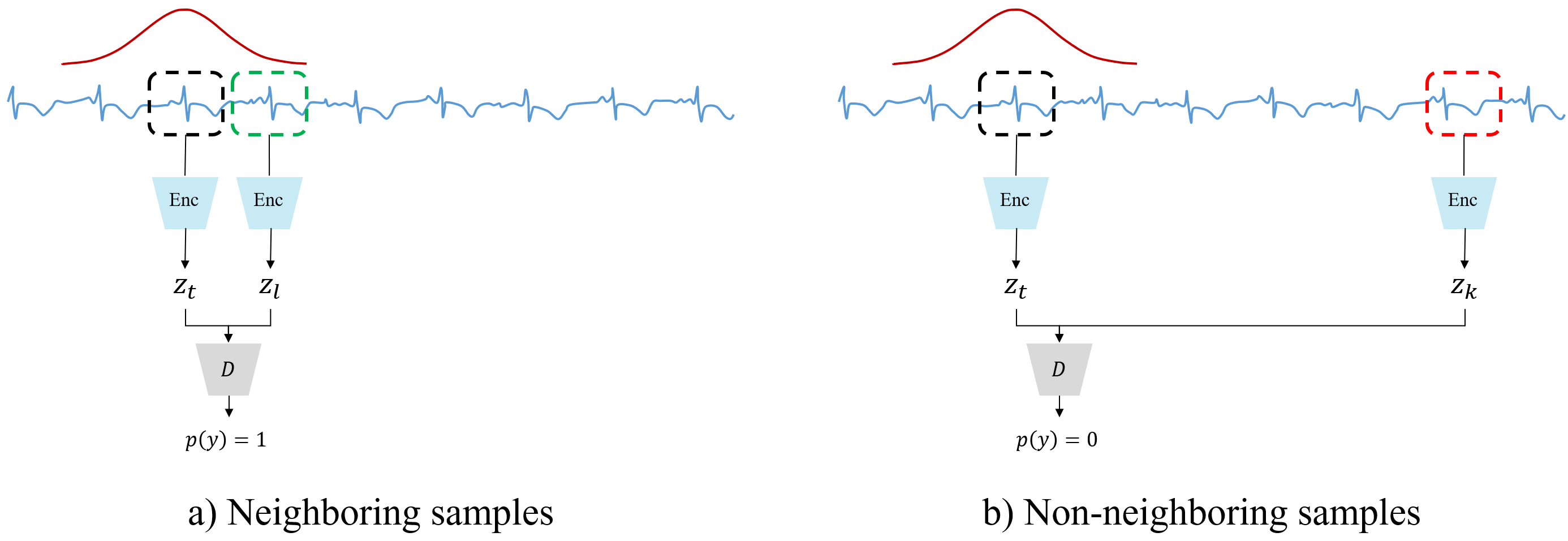}
    \caption{Overview of the TNC framework components. \textit{Enc}, $z$, and $D$ denote an encoder, representation, and a discriminator, where the discriminator is a shallow neural network for binary classification. TNC encodes the distinguishable distributions between neighboring samples and non-neighboring samples by training the discriminator predict 1 for representations of neighboring samples (denoted as $z_t$ and $z_l$) and 0 for representations of non-neighboring samples (denoted as $z_t$ and $z_k$).}
    \label{fig:tnc}       % Give a unique label
\end{figure}

Simplified illustrations of the introduced SSL methods are presented in Fig. \ref{fig:ssl_methods}. It should be noted that time series or spectrogram can be used as an input instead of an image while the entire methodological framework remains intact. Also, an overview of the TNC framework is presented in Fig. \ref{fig:tnc}. While the figure shows time series as input, it should be noted that spectrograms can be used as input instead of time series, which is the case in our study.

In this study, VNIbCReg (Variance Neighboring-Invariance better-Covariance Regularization) is proposed. It can be viewed as a combination of a VICReg-style method and TNC with neighboring-invariance and better-covariance. It is shown to be effective in learning quality representations on non-stationary time series. In particular, VIbCReg is used instead of VICReg as VIbCReg was shown to be more effective than VICReg on time series. The evaluation is conducted by the linear evaluation and the fine-tuning evaluation on small subsets of the training set. As for the dataset, non-stationary seismic signal time series is used. A detailed description about the dataset is presented in the following section.

The source code is available at \url{https://github.com/ML4ITS/Aknes_clf}. The used dataset, called Åknes dataset, is included in the source code, but can also be found at \url{https://bit.ly/3EZkthW}. For the Åknes dataset, events were classified by visual inspection of a large number of waveforms recorded in different years, and proper labelling was ensured by cross-checking with a reviewed seismic bulletin \cite{seismic_bulletins}. Therefore, this dataset is encouraged to be used as a benchmark dataset so that different classification methods can be developed and fairly compared to each other.

\section{Background on the Dataset and Dataset Description}

\begin{figure}[htb]
    \centering
    \includegraphics[width=0.7\textwidth]{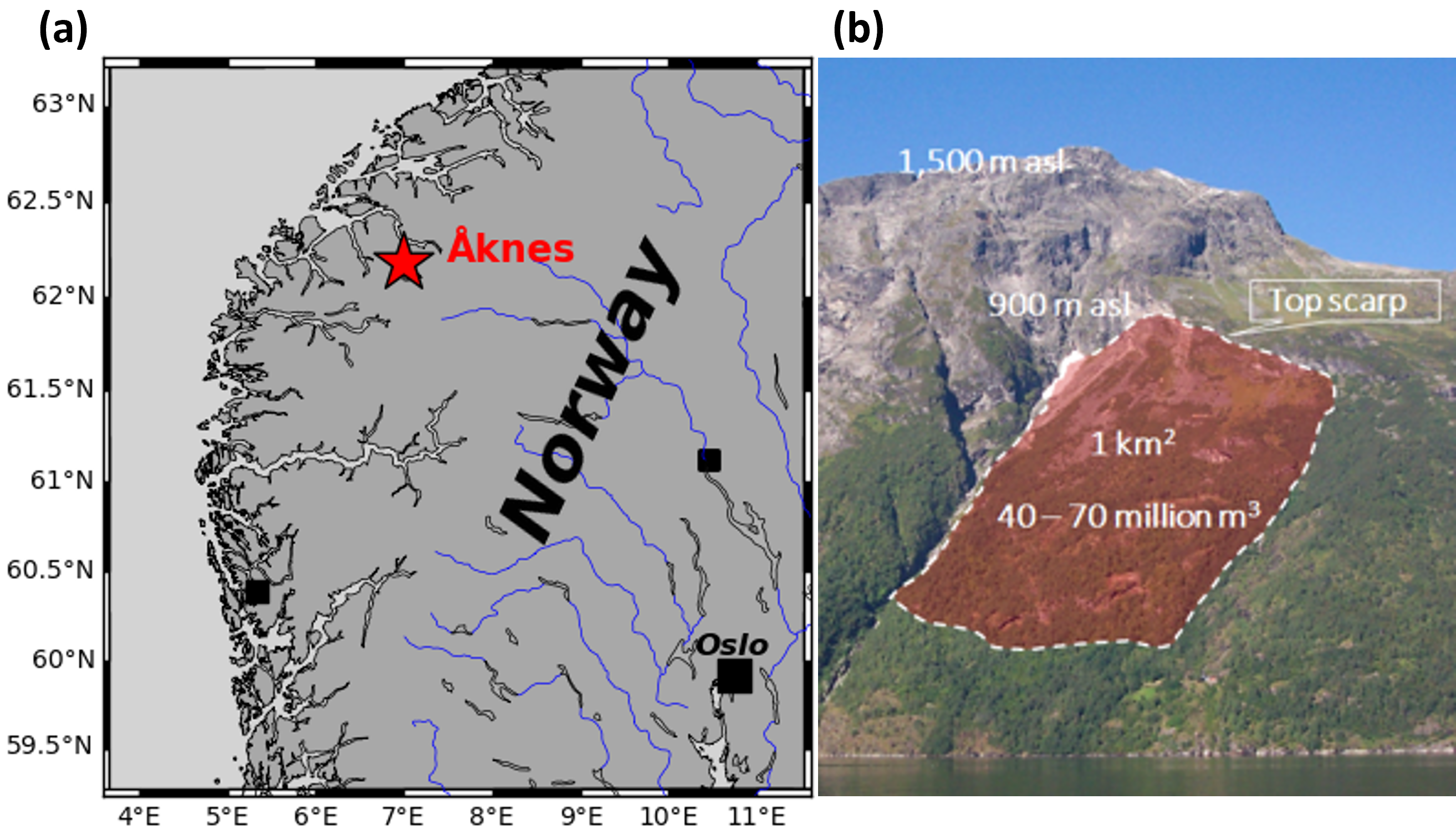}
    \caption{(a) Location of the Åknes rockslope in Norway. (b) Photo of the slope taken from the opposite site of the fjord \cite{roth2010}. The unstable area is highlighted by the red shade.}
    \label{fig:mapAaknes}       % Give a unique label
\end{figure}

The data is obtained from an Åknes rock slope. The Åknes rock slope is located south of Stranda in Western Norway, see Figure \ref{fig:mapAaknes}, by the fjord going in to Hellesylt and Geiranger.
The steep slope goes from the fjord up to nearly $1500$ m altitude. The unstable rock slope is limited by strike-slip faults to the North-East and South-West, and a fracture/backscarp as the upper limit. The upper fracture has lately been widening at about 2 cm per year, and this has caused concern about rock fall geohazards.

The risk-prone rock slope volume is up to 50 million m$^3$. Even though three dimensional numerical modeling of the slope has been conducted \cite{gharti2012application}, it has shown difficult to map the subsurface properties of the slope due to large heterogeneities in the subsurface seismic velocity model. There are hence large uncertainties associated with the unstable volume, and how it will collapse into the fjord. But even with a much smaller volume, rock collapse would create a tsunami with severe impacts on the local communities \cite{harbitz2014rockslide}, particularly considering its location in the narrow Norwegian fjords. The natural beauty of this fjord region has made it a UNESCO World Heritage Site. The area attracts nearly a million tourists per year and about 200 cruise ships pass by the Åknes rock slope every year. 

Because of the widening of the upper crack at Åknes, the rock slope has been extensively monitored for several years. In these efforts one has gathered different data sources such as surface crack displacements sensors \cite{nordvik2009statistical} and the displacements connections to meteorological data \cite{groneng2011meteorological}, InSAR data \cite{bardi2016space} and seismic geophone data \cite{roth2006seismic}.
Here, the focus is mainly on the geophone data. 

Altogether, there are eight geophones in the slope, each of them with three-component sensors. The geophone data passively register seismic activity at Åknes. The data are streamed to a computer server and this hence provides a continuous-time monitoring system for detecting potential rock fall or movements on the rock slope. When there is a seismic signal above a pre-specified amplitude threshold, a window of 16 seconds of geophone data is stored. Based on expert opinion, every event is then classified into one of eight classes, as described in Table~\ref{tab:class_types}. Each of these classes present specific characteristics and are likely associated with different physical mechanisms either directly involved in the movement of the rockslope or occurring farther away. Discussing these processes is out of the scope of this work, but more details can be found in \cite{nadege_paper}.
There is by now a rich database of geophone observations and associated rock fall class interpretation for various seismic events at Åknes \cite{nadege_paper}. This provides an excellent test case to understand and try out new machine learning methods for classifying rock slope events, and by building a reliable encoder and classifier, one can achieve automatic classification of such seismic geophone data events in the future. This would provide a low-cost and highly valuable decision support tool in the context of geohazard warning systems. 

\begin{table}[htb]
\centering
\caption{Class types with the corresponding numbers of samples in the dataset. The dataset is available within the source code. The data has been acquired over many years. The data with the valid classes (\textit{i.e.,} Noise to Spike) had been acquired from 2007 to 2020 and the data with the Unlabeled class was acquired throughout 2021.}
\label{tab:class_types}
\begin{tabular}{lc}
\hline\noalign{\smallskip}
class name               & a number of samples \\
\noalign{\smallskip}\hline\noalign{\smallskip}
Noise                     & 8                   \\
Regional                  & 292                 \\
%Rockfall short            & 207                 \\
%Rockfall wide             & 215                 \\
Rockfall                 & 215                 \\
Slope high-frequency (HF) & 448                 \\
Slope low-frequency (LF)  & 218                 \\
Slope multi               & 207                 \\
Slope tremor              & 212                 \\
Spike                     & 218                 \\
Unlabeled                 & 1611               \\
\noalign{\smallskip}\hline
\end{tabular}
\end{table}

\begin{figure}[htb]
    \centering
    \includegraphics[width=1\textwidth]{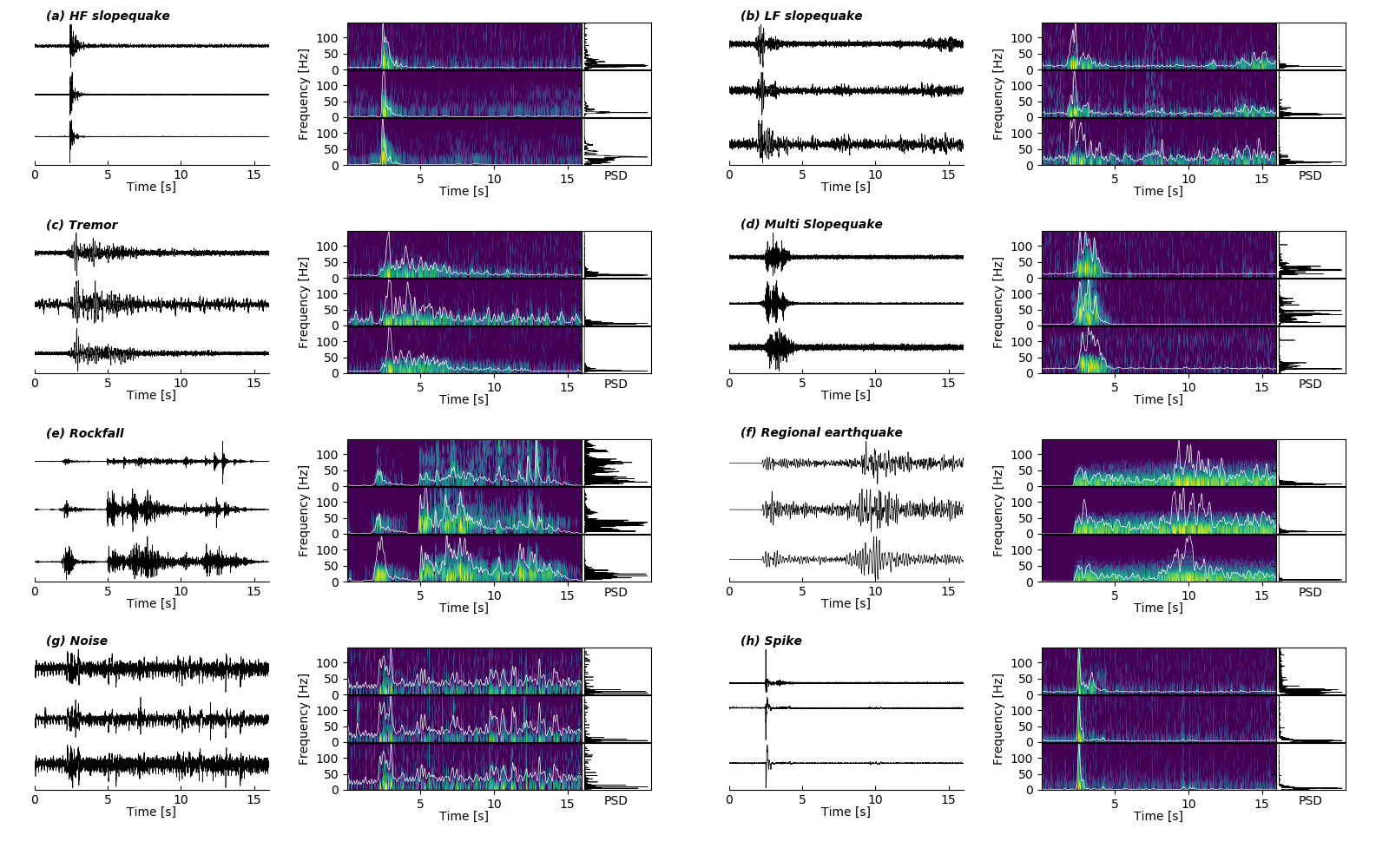}
    \caption{Example time series (left) associated with their spectrograms (right) for each class (a-g). Vertical component records of a same event at three chosen geophones are shown for each class. White lines on spectograms delineate signal envelopes while plots on the right show their Power Spectral Density (PSD).}
    \label{fig:example_spectrograms}       % Give a unique label
\end{figure}

Examples of 16 seconds geophone data recordings are shown in Fig.~\ref{fig:example_spectrograms}. The sampling frequency is $1000$ Hz, so the time series data consist of length-$16$ $000$ vectors for each of the eight geophones with three $(x,y,z)$ components (24 time series in total). These data represent non-stationary time series, where the amplitude and frequency content clearly change over the $16$ second time interval. It is not obvious how to summarize such an image by simple statistics such as mean, variance, correlation, dominating frequency, or other measures. 

A common technique of analysing such data is to compute spectrograms \cite{cohen1989time}. This essentially entails taking a rolling window short-time Fourier transformation of the data. The magnitudes for each frequency and time will then indicate the core frequency contents of the signal as a function of the $16$ second time interval. Figure \ref{fig:example_spectrograms} shows example spectrograms associated with the time series. 

There are apparent differences in the signals. The most obvious one is probably the signal duration which can last less than 3~seconds (Fig.~\ref{fig:example_spectrograms}a, b, d, h) or, in the contrary, last longer or even span the full duration of the record (Fig.~\ref{fig:example_spectrograms}c, e, f, g). The shape of the signals, represented by their envelopes, is also an important characteristic and displays either a main peak of energy (e.g.~Fig.~\ref{fig:example_spectrograms}a, h) or several peaks of energy distributed over time (e.g.~Fig.~\ref{fig:example_spectrograms}c, e, f). Lastly, the frequency content of the signals exhibit clear differences from a type of event to another, with some classes reaching high frequencies up to 80~Hz or even more (e.g.~Fig.~\ref{fig:example_spectrograms}a, d, e, h) while the other are confined to lower frequencies ($<20$~Hz, e.g.~Fig.~\ref{fig:example_spectrograms}b, c, f). Moreover, variability in the time series, and consequently in the spectrograms, can be observed depending on the sensor which recorded the event. This is particularly visible for the rockfall event (Fig.~\ref{fig:example_spectrograms}e) where the energy bandwidth decreases for the records from top to bottom. In this example, the relative amplitudes of the different bursts in the time series are also variable. Such variability from a geophone to another is mostly due to the event source location and its distance to the recording stations. Seismic waves are indeed affected by the properties of the medium in which they propagate.

\section{Proposed Method: VNIbCReg}

As mentioned earlier, VNIbCReg can be viewed as VIbCReg with TNC with neighboring-invariance and better-covariance. VIbCReg is effective in representation learning because samples with different classes can be well separated in the representation space. Despite the effectiveness of VIbCReg, it was not designed to encode temporal transition in the (signal) representation. The necessity of encoding the temporal transition can be observed given several samples from our dataset in Fig. \ref{fig:example_spectrograms}. To compensate for the lack of ability of encoding the temporal transition in VIbCReg, TNC is adopted into VIbCReg, resulting in VNIbCReg.

\begin{figure}[ht]
    \centering
    \includegraphics[width=0.7\textwidth]{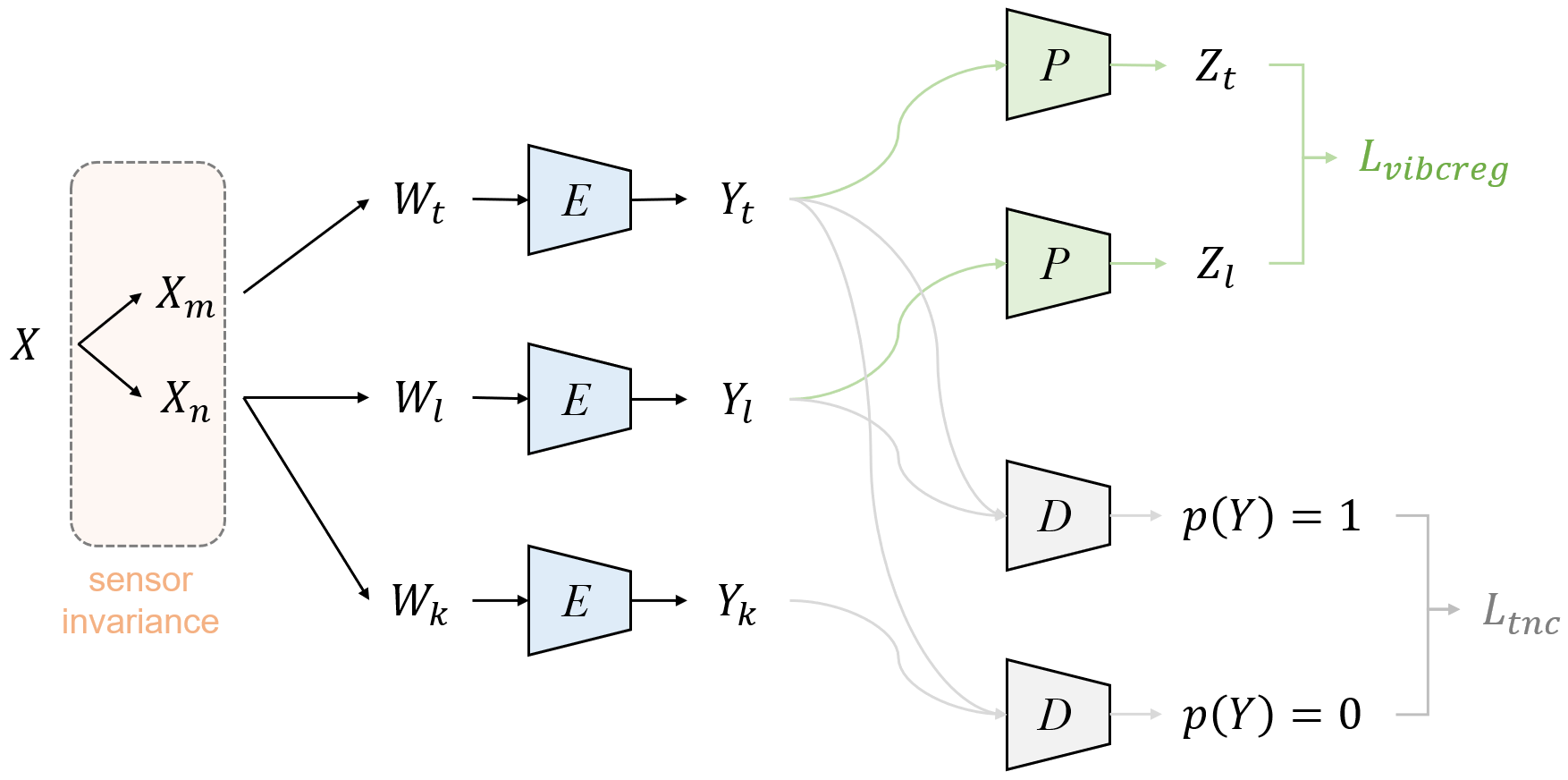}
    \caption{Overall framework of VNIbCReg.}
    \label{fig:vnibcreg}
\end{figure}

\begin{figure}[ht]
    \centering
    \includegraphics[width=0.7\textwidth]{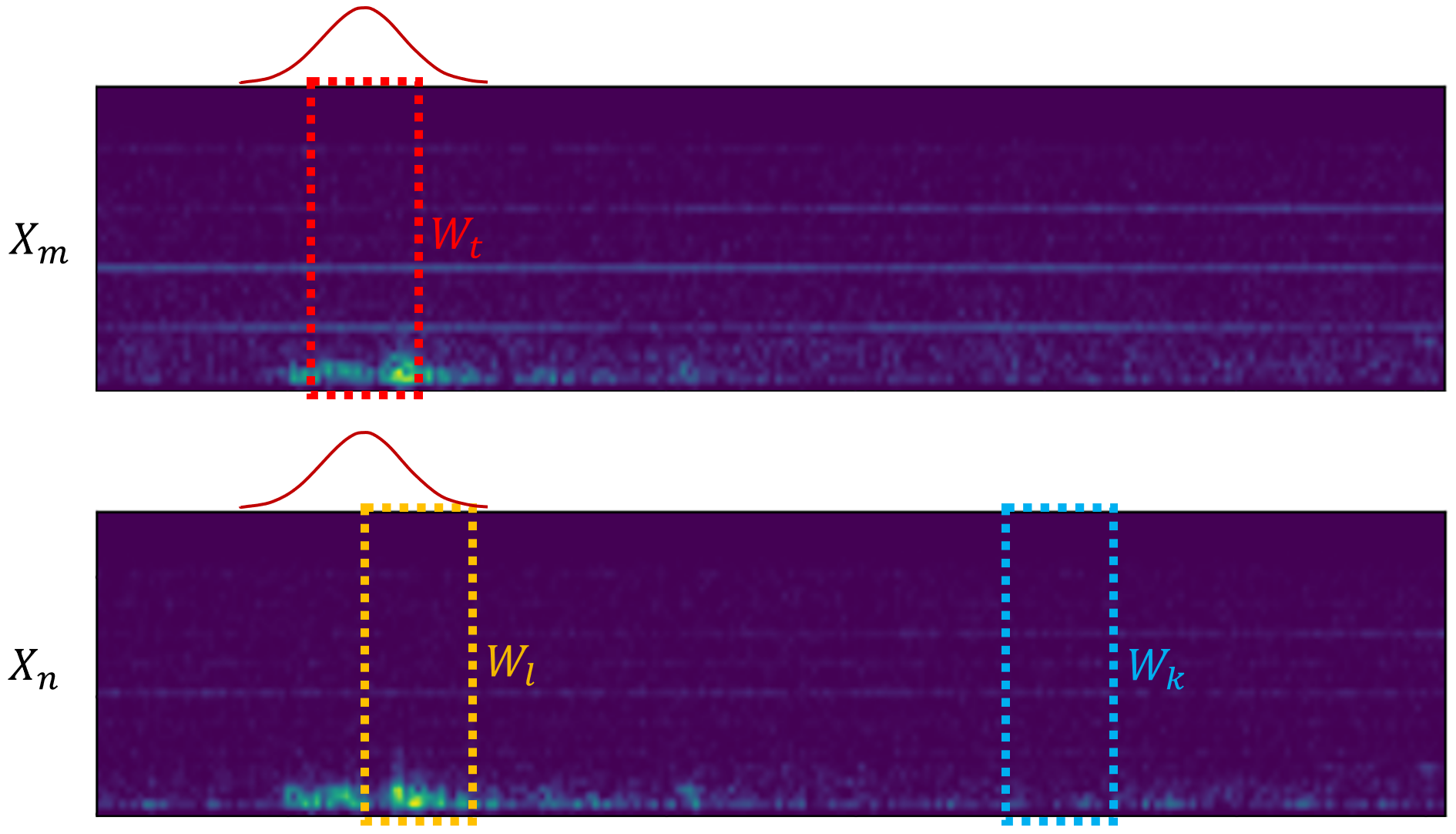}
    \caption{Illustration of $W_t$, $W_l$, and $W_k$. Each denotes a reference window, a neighboring window, and a non-neighboring window. The red curve of a normal distribution shape represents the neighborhood.}
    \label{fig:nc_spectrogram}
\end{figure}

The overall framework of VNIbCReg is presented in Fig. \ref{fig:vnibcreg}. Notation is clarified next, along with the required loss functions.  Here, $X$ denotes an input of the spectrograms where $X \in \mathbb{R}^{B \times 24  \times H \times W}$ ($B$: batch, $H$: height, and $W$: width), and $m$ and $n$ denote randomly-selected indices for a single time series among the 24 time series. Therefore, $X_m$ and $X_n$ have dimension $\mathbb{B}^{B \times H \times W}$. A cropped window is denoted by $W$, where $W_t$ and $W_l$ are in the same neighborhood while $W_t$ and $W_k$ are non-neighboring. It should be noted that $W_t$, $W_l$, and $W_k$ are randomly chosen while keeping their spatial relations to each other. An illustration of $W_t$, $W_l$, and $W_k$ is presented in Fig. \ref{fig:nc_spectrogram}. Further, $E$, $P$, and $D$ denote an encoder, the projector in VIbCReg, and the discriminator in TNC, respectively. Loss functions are denoted by $L$. Note that an iterative normalization layer in VIbCReg is omitted for simplicity in the figure and the same coloring for models such as $E$, $P$, and $D$ represents the shared weights between the same colored models. As for the discriminator's input, two $Y$-s are concatenated and used as input. As for $Y$ and $Z$ in the loss functions, the following notation is used: $Y = [y_1, \cdots, y_B]^T \in \mathbb{R}^{B \times F_y}$ and $Z = [z_1, \cdots, z_B]^T \in \mathbb{R}^{B \times F_z}$, where $F_y$ and $F_z$ denote feature size of $Y$ and $Z$, respectively.

The additive loss is used for the two parts:
\begin{equation} \label{eq:vnibcreg}
    L_{vnibcreg} = L_{vibcreg} + L_{tnc}.
\end{equation}
Here, the first part on the right side of Eqn. \ref{eq:vnibcreg}
consists of 
\begingroup
\allowdisplaybreaks
\begin{align}
\label{eq:vibcreg}
L_{vibcreg} =& \lambda s(Z_t, Z_l) + \mu \{v(Z_t) + v(Z_l)\} + \nu \{c(Z_t) + c(Z_l)\}, \\
&s(Z_t, Z_l) = \frac{1}{B} \sum_{b=1}^B \| Z_{t_b} - Z_{l_b} \|_2^2, \\
&v(Z) = \frac{1}{F_z} \sum_{f=1}^{F_z} \mathrm{ReLU}\left( \gamma - \sqrt{\mathrm{Var}\left( Z_f \right) + \epsilon} \right), \\
&c(Z) = \frac{1}{F_z^2} \sum_{i \neq j} C(Z)^2_{i, j}, \\
&C(Z) = \left( \frac{Z - \bar{Z}}{\| Z - \bar{Z} \|_2} \right)^T \left( \frac{Z - \bar{Z}}{\| Z - \bar{Z} \|_2} \right) \:\: \mathrm{where} \:\: \bar{Z}=\frac{1}{B}\sum_{b=1}^{B}{Z_b}.
\end{align}
\endgroup
Here, $s(..)$ denotes the \textit{invariance} term, $v(..)$ \textit{variance} term and $c(..)$ the \textit{covariance} term. Moreover, $\mathrm{Var}$ denotes a variance estimator, $\gamma$ is a target value for the standard deviation, fixed to $1$ as in the original implementation, $\epsilon$ is a small scalar (\textit{i.e.,} $0.0001$) to prevent numerical instability, $\sum_{i \neq j}$ denotes summation of off-diagonal terms in a 2-dimensional matrix, and $\lambda$, $\mu$, and $\nu$ are hyper-parameters to control the importance of each term. It should be noted that the notation for $L_{vibcreg}$ largely follows that of the original paper. 

The second term on the right in Eqn. \eqref{eq:vnibcreg} is defined as follows:
\begingroup
\allowdisplaybreaks
\begin{align}
\label{eq:tnc}
L_{tnc} =& \rho \left( \{ -\log{(D(Y_t, Y_l))} + \log{(1 - D(Y_t, Y_k))} \} + \{ (1 - c_p(Y_t, Y_l))^2 + c_n(Y_t, Y_k)^2 \} \right), \\
&c_p(Y_t, Y_l) = \frac{1}{F_y} \sum_{i=j}{ \left( \frac{Y_t - \bar{Y_t}}{\| Y_t - \bar{Y_t} \|} \right)^T \left( \frac{Y_l - \bar{Y_l}}{\| Y_l - \bar{Y_l} \|} \right) \:\: \mathrm{where} \:\: \bar{Y}=\frac{1}{B}\sum_{b=1}^{B}{Y_b}}, \\
&c_n(Y_t, Y_k) = \frac{1}{F_y} \sum_{i=j}{ \left( \frac{Y_t - \bar{Y_t}}{\| Y_t - \bar{Y_t} \|} \right)^T \left( \frac{Y_k - \bar{Y_k}}{\| Y_k - \bar{Y_k} \|} \right) }.
\end{align}
\endgroup
The original TNC loss function is defined as $-\log{(D(Y_t, Y_l))} + \log{(1 - D(Y_t, Y_k))}$ given that its regularization weighting hyper-parameter is set to zero. In Eqn. \eqref{eq:tnc}, there is an additional term, $(1 - c_p(Y_t, Y_l))^2 + c_n(Y_t, Y_k)^2$, which is proposed added. This addition improves quality of representation learning on the top of the original TNC loss by correlating $Y_t$ and $Y_l$ and de-correlating $Y_t$ and $Y_k$. Here, $\sum_{i = j}$ denotes summation of diagonal terms in a 2-dimensional matrix and $\rho$ is a hyper-parameter for weighting $L_{tnc}$. In the experiments, $\lambda$, $\mu$, $\nu$, and $\rho$ are empirically set to 10, 10, 10, and 13, respectively. 

Lastly, a quirky component in the VNIbCReg framework is the \textit{sensor invariance} which is proposed for the following reason: A single sample consists of the 24 time series in our dataset. Although the 24 time series are different to some extent due to the sensors' location, functionality, and axial directions, they are semantically the same in a sense that they belong to the same class. The sensor invariance is what allows representations of the spectrograms of the different time series within the same sample to be pulled together in the embedding space by both VIbCReg and TNC. 

After pre-training by a SSL method, only the encoder is typically kept while discarding the rest of the parts such as the projector and the discriminator. Then, the encoder can be used for various downstream tasks of classification with some fine-tuning.

\section{Results and Discussion}
\label{sect:results}

\subsection{Implementation Details}
\label{subsect:implementation_details}

\paragraph{Training and Test Datasets}
For the naive supervised learning and the linear evaluation, the dataset is split by stratified random sampling into 80\% for training and 20\%  for testing. For the fine-tuning evaluation, the dataset is split into $n$\% (where $n$ is specified in the fine-tuning result table) and 20\% of the dataset for training and test datasets, respectively. For the naive supervised learning, the linear evaluation, and the fine-tuning evaluation, samples with valid classes (\textit{i.e.,} noise, regional, rockfall, slope HF, slope LF, slope tremor, slope multi, and spike) are used as a dataset. For the SSL, its dataset consists of all the samples including the unlabeled-class samples. Given the randomness of the dataset split, all the experimental cases are run three times with different random seeds. 

\paragraph{Preprocessing}
First, a time series input is converted into a spectrogram using a spectrogram function from \textit{SciPy} with a 0.08 s-long sliding window with a 12.5\% overlap \cite{2020SciPy-NMeth}. Then, it is scaled by $\log$ and min-max scaling, and resized such that the height is adjusted to 128 and the width is adjusted proportionally to the increase ratio of the height. 

The preprocessing used in our experiments is somewhat different from that of \cite{nadege_paper} in which spectrograms are computed using a 1 sec sliding window with a 95\% overlap, and then converted into an RGB image with the size of $(3 \times 224 \times 224)$. The sliding window length and overlap give monotonous spectrograms \textit{i.e.,} low-resolution spectrogram, and it leads to an overfitting which give lower accuracy in our experiments. Also, the spectrograms do not have to be converted into square-shaped RGB images. Spectrograms originally have dimension of $(1 \times H \times W)$ as 1 channel real-valued matrix, where $H$ and $W$ denote height and width, respectively. Thus, a naive form of spectrograms can be already viewed as an 1-dimensional real-valued image data. In addition, because a convolutional layer can process rectangle image shapes as well as square shapes, the rectangle shapes of the naive form of spectrograms can be processed by a CNN model without squishing the rectangular shape into a square, which can actually lead to some information loss and performance drop. 

\paragraph{Augmentation} In our experiments, there are two types of augmentations: Neighboring Crop (NC) and Random Crop (RC). The NC takes a spectrogram and outputs three different cropped windows of the spectrogram, $W_t$, $W_l$, and $W_k$, as shown in Fig. \ref{fig:vnibcreg}. In contrast, the RC outputs three different cropped windows of the spectrogram. But the cropped windows in the RC do not have any spatial relation between each other as the three cropped windows are selected at random. In our experiments, height of the crop is set to the same height as the input spectrogram and the width of the crop is set to 10\% of width of the input spectrogram for both the NC and the RC. Although additional augmentation methods can likely gain extra performance improvement, no other method is used here to clearly compare the performance between the RC and the NC in the linear and fine-tuning evaluations.

\paragraph{Architecture}
%In our experiments, AlexNet and ResNet34 are compared. AlexNet is the standard CNN model used in the relevant previous studies \cite{nadege_paper} and ResNet is a very promising CNN model. ResNet34 is specifically chosen due to its representation size as 512 which is the same as in VIbCReg and has bigger model capability than ResNet18. It should be noted that the numbers of parameters for AlexNet and ResNet34 are around 57 million and 21 million, respectively. Hence, ResNet34 is almost three times \textit{lighter} than AlexNet. The original implementation of AlexNet and ResNet takes RGB images (\textit{i.e.,} $C \times H \times W$, where $C$ is 3) as input. Therefore, the input channel size is 3 for the first convolutional layers in both models. In our experiments, the input spectrogram has a dimension of $(1 \times H \times W)$, therefore, the first convolutional layers in both models are modified to receive an input with one channel. Except that, both model architectures remain the same as in \cite{nadege_paper}. The same holds for the projector and the discriminator.

In our experiments, ResNet34 is used. 
%AlexNet is the standard CNN model used in the relevant previous studies \cite{nadege_paper} and ResNet is a very promising CNN model. ResNet34 is specifically chosen due to its representation size as 512 which is the same as in VIbCReg and has bigger model capability than ResNet18. It should be noted that the numbers of parameters for AlexNet and ResNet34 are around 57 million and 21 million, respectively. Hence, ResNet34 is almost three times \textit{lighter} than AlexNet. 
The original implementation \cite{nadege_paper} takes spectrograms converted into RGB images (\textit{i.e.,} $C \times H \times W$, where $C$ is 3) as input. Therefore, the input channel size is 3 for the first convolutional layers in both models. In our experiments, the input spectrogram has a dimension of $(1 \times H \times W)$, therefore, the first convolutional layers in both models are modified to receive an input with one channel. Except that, both model architectures remain the same as in \cite{nadege_paper}. The same holds for the projector and the discriminator.

\paragraph{Optimizer}
Adam \cite{kingma2014adam} is used with a cosine learning rate scheduler. Its initial learning rate for the encoder and the classifier is set to 0.001 for the SSL and the linear evaluation. For the fine-tuning evaluation on the SSL methods, the initial learning rates are set to 0.0005 and 0.001 for the encoder and the classifier, respectively. The batch size is 128 for the SSL and 64 for naive supervised learning, the linear evaluation, and the fine-tuning evaluation. A number of epochs is 300 for the self-supervised learning and 100 for the linear evaluation and the fine-tuning evaluation. The used deep learning library is \textit{PyTorch} \cite{NEURIPS2019_9015}.

\subsection{Results}

\paragraph{Linear and Fine-tuning Evaluations on SSL Methods}

The linear and fine-tuning evaluations are conducted on VIbCReg, VNIbCReg, and intermediate variants between VIbCReg and VNIbCReg. The evaluation on the intermediate variants are conducted to find out how much contribution each component that is added on VIbCReg to form VNIbCReg makes. 

The linear evaluation results are presented in Table \ref{tab:LE}. VIbCReg does not perform so well here, meaning that the learned representations by VIbCReg are not so linearly-separable by class. By introducing main components from the original TNC (\textit{i.e.,} NC and $L_{tnc}^{\dagger}$), a significant improvement is achieved in the linear evaluation. This indicates that the ability of encoding the temporal transition improves quality of learned representations especially for non-stationary time series with apparent temporal transition. 

The effectiveness of $L_{tnc}$ is shown by further improving the performance. The fine-tuning evaluation result is presented in Table \ref{tab:FT}. The performance ranking order of the fine-tuning evaluation between VIbCReg, VNIbCReg, and the intermediate variants remains the same as in the linear evaluation. One of the noticeable points in the result is the high test accuracy in a small dataset regime (\textit{i.e.,} $n$ of 5\% and 10\%) for models that are pretrained by a SSL method. Especially, VNIbCReg results in the highest test accuracy in the small dataset regime, indicating that the model is well generalized by VNIbCReg. As the geophone sensor data is collected unlabeled, there is naturally a large unlabeled dataset while the labeled dataset is much smaller because proper labeling can only be done manually by an expert. Given that circumstance, VNIbCReg is expected to provide robust performance improvement when a large amount of unlabeled dataset is utilized.

\begin{table}[ht]
\centering
\caption{Linear evaluation result. RC and NC refer to the random crop and neighboring crop, respectively. $L_{tnc}^{\dagger}$ denotes the original TNC loss and $L_{tnc}$ denotes the modified TNC loss proposed in this work. o and x denote used and not-used. Note that VNIbCReg corresponds to the based SSL method of VIbCReg with the NC and $L_{tnc}$. RandInit refers to a case where an encoder is randomly-initialized and frozen without any pre-training and used for the linear evaluation.}
\label{tab:LE}
\begin{tabular}{lccccl}
\hline\noalign{\smallskip}

\Centerstack{Base SSL\\method} & \Centerstack{Augmentation \\ for cropping} & $L_{tnc}^{\dagger}$ & $L_{tnc}$ & Test acc. & Remarks \\
%  & \begin{tabular}[c]{@{}l@{}}augmentation \\ for cropping\end{tabular} & $L_{tnc}^{\dagger}$ & $L_{tnc}$ & Test acc. \\

\noalign{\smallskip}\hline\noalign{\smallskip}
RandInit & x  & x & x & 0.642 (0.026) &          \\
TNC & NC & o & x & 0.499 (0.011) & = naive TNC \\
VIbCReg  & RC & x & x & 0.476 (0.02) & = naive VIbCReg            \\
VIbCReg  & NC & x & x & 0.578 (0.005) &          \\
VIbCReg  & NC & o & x & \underline{0.698 (0.023)} &  \\
VIbCReg  & NC & x & o & \textbf{0.716 (0.025)} & \textbf{= VNIbCReg} \\
\noalign{\smallskip}\hline
\end{tabular}
\end{table}

\begin{table}[htb]
\centering
\caption{Fine-tuning evaluation result. The notation is the same as in the linear evaluation result. RandInit refers to a case where an encoder is randomly-initialized and frozen without any pre-training and used for the fine-tuning evaluation (\textit{i.e.,} naive supervised learning). Note that the accuracy is much higher with a small labeled dataset when pretrained by VNIbCReg.}
\label{tab:FT}
\begin{tabular}{lcccccc}
\hline\noalign{\smallskip}

\Centerstack{Base SSL\\method} & \Centerstack{Augmentation \\ for cropping} & $L_{tnc}^{\dagger}$ & $L_{tnc}$ & \multicolumn{3}{c}{Test acc. (fine-tuned on $n \%$ of the dataset)}\\ 
\noalign{\smallskip}\cline{5-7}\noalign{\smallskip}
\multirow{2}{*}{} & & & & $n=5(\%$) & $n=10(\%)$ & $n=80(\%)$ \\
  
\noalign{\smallskip}\hline\noalign{\smallskip}
RandInit & x  & x & x & 0.185 (0.036)          & 0.528 (0.038)         & \textbf{0.928 (0.005)} \\
TNC & NC & o & x & 0.5 (0.029) & 0.715 (0.033) & 0.915 (0.01) \\
VIbCReg  & RC & x & x & 0.504 (0.036)          & 0.628 (0.024)         & 0.912 (0.002)          \\
VIbCReg  & NC & x & x & 0.619 (0.036)          & 0.747 (0.016)         & 0.912 (0.003)          \\
VIbCReg  & NC & o & x & {\underline{0.642 (0.019)}}    & {\underline{0.799 (0.029)}}   & {\underline{0.92 (0.002)}}     \\
VIbCReg  & NC & x & o & \textbf{0.717 (0.037)} & \textbf{0.824 (0.03)} & {\underline{0.919 (0.005)}}  \\
\noalign{\smallskip}\hline
\end{tabular}
\end{table}

\section{Conclusion}
Although SSL methods such as VICReg on images and VIbCReg on time series were successful on their benchmark datasets, they were not designed to be able to consider non-stationarity. VNIbCReg which is a combination of VIbCReg and TNC is shown to be effective in learning representations of non-stationary time series. Given its effectiveness, a further study can be conducted on more non-stationary time series datasets to verify its general SSL capability.

\section*{Acknowledgments}
We acknowledge support from the Norwegian Research Council project ML4ITS grant 312062 and the SFI Centre for Geophysical Forecasting grant 309960. 

%Bibliography
\bibliographystyle{unsrt}  
\bibliography{references}

\begin{thebibliography}{10}

\bibitem{lee2022ensemble}
Daesoo Lee, Erlend Aune, Nad{\`e}ge Langet, and Jo~Eidsvik.
\newblock Ensemble and self-supervised learning for improved classification of
  seismic signals from the {\aa}knes rockslope.
\newblock {\em Mathematical Geosciences}, pages 1--24, 2022.

\bibitem{he2020momentum}
Kaiming He, Haoqi Fan, Yuxin Wu, Saining Xie, and Ross Girshick.
\newblock Momentum contrast for unsupervised visual representation learning.
\newblock In {\em Proceedings of the IEEE/CVF conference on computer vision and
  pattern recognition}, pages 9729--9738, 2020.

\bibitem{chen2020simple}
Ting Chen, Simon Kornblith, Mohammad Norouzi, and Geoffrey Hinton.
\newblock A simple framework for contrastive learning of visual
  representations.
\newblock In {\em International conference on machine learning}, pages
  1597--1607. PMLR, 2020.

\bibitem{grill2020bootstrap}
Jean-Bastien Grill, Florian Strub, Florent Altch{\'e}, Corentin Tallec, Pierre
  Richemond, Elena Buchatskaya, Carl Doersch, Bernardo Avila~Pires, Zhaohan
  Guo, Mohammad Gheshlaghi~Azar, et~al.
\newblock Bootstrap your own latent-a new approach to self-supervised learning.
\newblock {\em Advances in Neural Information Processing Systems},
  33:21271--21284, 2020.

\bibitem{chen2021exploring}
Xinlei Chen and Kaiming He.
\newblock Exploring simple siamese representation learning.
\newblock In {\em Proceedings of the IEEE/CVF Conference on Computer Vision and
  Pattern Recognition}, pages 15750--15758, 2021.

\bibitem{zbontar2021barlow}
Jure Zbontar, Li~Jing, Ishan Misra, Yann LeCun, and St{\'e}phane Deny.
\newblock Barlow twins: Self-supervised learning via redundancy reduction.
\newblock In {\em International Conference on Machine Learning}, pages
  12310--12320. PMLR, 2021.

\bibitem{bardes2021vicreg}
Adrien Bardes, Jean Ponce, and Yann LeCun.
\newblock Vicreg: Variance-invariance-covariance regularization for
  self-supervised learning.
\newblock {\em arXiv preprint arXiv:2105.04906}, 2021.

\bibitem{lee2021vibcreg}
Daesoo Lee and Erlend Aune.
\newblock {VIbCReg: Variance-Invariance-better-Covariance Regularization for
  Self-Supervised Learning on Time Series}.
\newblock {\em arXiv preprint arXiv:2109.00783}, 2021.

\bibitem{tonekaboni2021unsupervised}
Sana Tonekaboni, Danny Eytan, and Anna Goldenberg.
\newblock Unsupervised representation learning for time series with temporal
  neighborhood coding.
\newblock {\em arXiv preprint arXiv:2106.00750}, 2021.

\bibitem{seismic_bulletins}
NORSAR.
\newblock {NORSAR: NORSAR Seismic Bulletins}.
\newblock \url{https://www.norsar.no/seismic-bulletins/}, 1971.
\newblock https://doi.org/10.21348/b.0001.

\bibitem{roth2010}
M.~Roth and L.H. Blikra.
\newblock Seismic monitoring of the unstable rock slope at Åknes, norway.
\newblock {\em Extended abstract, Open Workshop within the frame of the EU FP7
  "SafeLand" Project February 24$^{th}$, 2010, Vienna (Austria)}, 2010.

\bibitem{gharti2012application}
Hom~Nath Gharti, Dimitri Komatitsch, Volker Oye, Roland Martin, and Jeroen
  Tromp.
\newblock Application of an elastoplastic spectral-element method to {3D} slope
  stability analysis.
\newblock {\em International Journal for Numerical Methods in Engineering},
  91(1):1--26, 2012.

\bibitem{harbitz2014rockslide}
CB~Harbitz, S~Glimsdal, F~L{\o}vholt, V~Kveldsvik, GK~Pedersen, and A~Jensen.
\newblock Rockslide tsunamis in complex fjords: From an unstable rock slope at
  {\aa}kerneset to tsunami risk in western {N}orway.
\newblock {\em Coastal engineering}, 88:101--122, 2014.

\bibitem{nordvik2009statistical}
Trond Nordvik and Erik Nyrnes.
\newblock Statistical analysis of surface displacements--an example from the
  {\aa}knes rockslide, western {N}orway.
\newblock {\em Natural Hazards and Earth System Sciences}, 9(3):713--724, 2009.

\bibitem{groneng2011meteorological}
Guro Gr{\o}neng, Hanne~H Christiansen, Bj{\o}rn Nilsen, and Lars~Harald Blikra.
\newblock Meteorological effects on seasonal displacements of the {\aa}knes
  rockslide, western {N}orway.
\newblock {\em Landslides}, 8(1):1--15, 2011.

\bibitem{bardi2016space}
Federica Bardi, Federico Raspini, Andrea Ciampalini, Lene Kristensen, Line
  Rouyet, Tom~Rune Lauknes, Regula Frauenfelder, and Nicola Casagli.
\newblock {Space-borne and ground-based InSAR data integration: the {\AA}knes
  test site}.
\newblock {\em Remote Sensing}, 8(3):237, 2016.

\bibitem{roth2006seismic}
Michael Roth, Michel Dietrich, Lars~H Blikra, and Isabelle Lecomte.
\newblock Seismic monitoring of the unstable rock slope site at {\aa}knes,
  norway.
\newblock In {\em Symposium on the Application of Geophysics to Engineering and
  Environmental Problems 2006}, pages 184--192. Society of Exploration
  Geophysicists, 2006.

\bibitem{nadege_paper}
N.~Langet and F.M.J. Silverberg.
\newblock {Automated classification of seismic signals recorded on the Åknes
  rockslope, Western Norway, using a Convolutional Neural Network}.
\newblock {\em Earth Surface Dynamics}, 2022.
\newblock (submitted).

\bibitem{cohen1989time}
Leon Cohen.
\newblock Time-frequency distributions-a review.
\newblock {\em Proceedings of the IEEE}, 77(7):941--981, 1989.

\bibitem{2020SciPy-NMeth}
Pauli Virtanen, Ralf Gommers, Travis~E. Oliphant, Matt Haberland, Tyler Reddy,
  David Cournapeau, Evgeni Burovski, Pearu Peterson, Warren Weckesser, Jonathan
  Bright, St{\'e}fan~J. {van der Walt}, Matthew Brett, Joshua Wilson, K.~Jarrod
  Millman, Nikolay Mayorov, Andrew R.~J. Nelson, Eric Jones, Robert Kern, Eric
  Larson, C~J Carey, {\.I}lhan Polat, Yu~Feng, Eric~W. Moore, Jake
  {VanderPlas}, Denis Laxalde, Josef Perktold, Robert Cimrman, Ian Henriksen,
  E.~A. Quintero, Charles~R. Harris, Anne~M. Archibald, Ant{\^o}nio~H. Ribeiro,
  Fabian Pedregosa, Paul {van Mulbregt}, and {SciPy 1.0 Contributors}.
\newblock {{SciPy} 1.0: Fundamental Algorithms for Scientific Computing in
  Python}.
\newblock {\em Nature Methods}, 17:261--272, 2020.

\bibitem{kingma2014adam}
Diederik~P Kingma and Jimmy Ba.
\newblock Adam: A method for stochastic optimization.
\newblock {\em arXiv preprint arXiv:1412.6980}, 2014.

\bibitem{NEURIPS2019_9015}
Adam Paszke, Sam Gross, Francisco Massa, Adam Lerer, James Bradbury, Gregory
  Chanan, Trevor Killeen, Zeming Lin, Natalia Gimelshein, Luca Antiga, Alban
  Desmaison, Andreas Kopf, Edward Yang, Zachary DeVito, Martin Raison, Alykhan
  Tejani, Sasank Chilamkurthy, Benoit Steiner, Lu~Fang, Junjie Bai, and Soumith
  Chintala.
\newblock {PyTorch: An Imperative Style, High-Performance Deep Learning
  Library}.
\newblock In H.~Wallach, H.~Larochelle, A.~Beygelzimer, F.~d\textquotesingle
  Alch\'{e}-Buc, E.~Fox, and R.~Garnett, editors, {\em Advances in Neural
  Information Processing Systems 32}, pages 8024--8035. Curran Associates,
  Inc., 2019.

\end{thebibliography}

\end{document}